%% file: main.tex
\documentclass[10pt,twocolumn,letterpaper]{article}

\usepackage[pagenumbers]{cvpr}
\input{macro}

\definecolor{cvprblue}{rgb}{0.21,0.49,0.74}
\usepackage[pagebackref,breaklinks,colorlinks,allcolors=cvprblue]{hyperref}

\title{ArtGen: Conditional Generative Modeling of Articulated Objects \\
in Arbitrary Part-Level States}

\author{
\textbf{Haowen Wang}$^{1}$ 
\and 
\textbf{Xiaoping Yuan}$^{2}$ 
\and 
\textbf{Fugang Zhang}$^{1}$ 
\and 
\textbf{Rui Jian}$^{1}$   
\and 
\textbf{Yuanwei Zhu}$^{1}$ 
\and 
\textbf{Xiuquan Qiao}$^{2}$ 
\and 
\textbf{Yakun Huang}$^{2}$   
\and
\\
$^1$Anhui University \hspace{5em} 
$^2$Beijing University of Posts and Telecommunications 
}

\begin{document}

\input{figures/fig0_intro}

\begin{abstract}
\input{sec/0_abstract}
\end{abstract}

\input{sec/1_introduction}

\input{figures/fig1_overview}

\input{sec/2_related_work}
\input{sec/3_methods}

\input{tables/tab1_sota}
\input{figures/fig3_sota}

\input{sec/4_experiments}

\input{sec/5_conclusion}

\newpage
{
    \small
    \bibliographystyle{ieeenat_fullname}
    \bibliography{main}
}

\end{document}

%% file: macro.tex
\usepackage{amsmath}   
\usepackage{multirow}  
\usepackage{enumitem}  

\def\methname{ArtGen}
\def\hiddenfeat{\textbf{Z}}

\newcommand{\mypara}[1]{\noindent\textbf{#1}}  
\usepackage{CJKutf8}

%% file: figures/fig0_intro.tex
\twocolumn[{
\renewcommand\twocolumn[1][]{#1}
\maketitle
\begin{center}
    \includegraphics[width=1.0\linewidth]{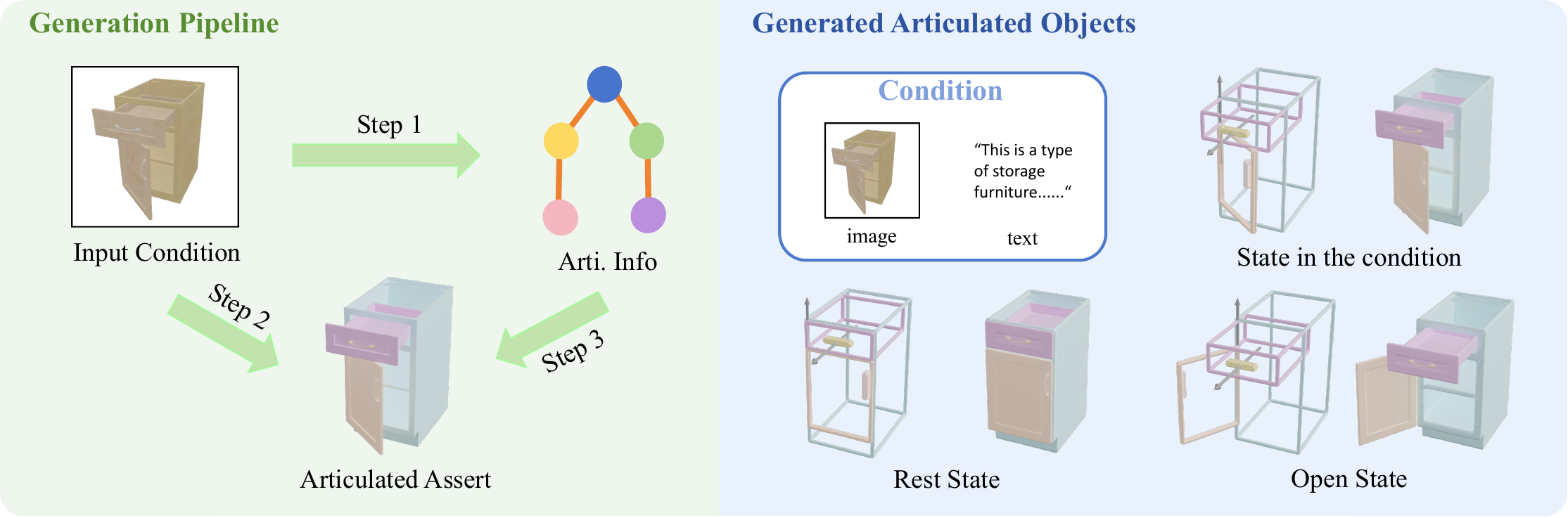}
    \captionsetup{type=figure}
    \caption{Given an input image or text description, our method generates articulated assets.
        Left: We design a pipeline for synthesizing articulated assets.
        Right: Our method generates geometric shapes and kinematic parameters for different states.
    }
    \label{fig:fig1_overview}
\end{center}
}]

%% file: sec/0_abstract.tex
Generating articulated assets is crucial for robotics, digital twins, and embodied intelligence. 
Existing generative models often rely on single-view inputs representing closed states, resulting in ambiguous or unrealistic kinematic structures due to the entanglement between geometric shape and joint dynamics. 
To address these challenges, we introduce ArtGen, a conditional diffusion-based framework capable of generating articulated 3D objects with accurate geometry and coherent kinematics from single-view images or text descriptions at arbitrary part-level states. 
Specifically, ArtGen employs cross-state Monte Carlo sampling to explicitly enforce global kinematic consistency, reducing structural-motion entanglement. 
Additionally, we integrate a Chain-of-Thought reasoning module to infer robust structural priors, such as part semantics, joint types, and connectivity, guiding a sparse-expert Diffusion Transformer to specialize in diverse kinematic interactions. 
Furthermore, a compositional 3D-VAE latent prior enhanced with local-global attention effectively captures fine-grained geometry and global part-level relationships.
Extensive experiments on the PartNet-Mobility benchmark demonstrate that ArtGen significantly outperforms state-of-the-art methods.

%% file: sec/1_introduction.tex
\section{Introduction}
\label{sec:intro}

Articulated objects such as kitchen cabinets, refrigerators, and storage drawers are ubiquitous in everyday environments and play an essential role in applications including robotics, augmented reality, and virtual reality~\cite{zhao2024tac, simeonov2022neural, DIFFTACTILE, guo2024extending, mangalam2024enhancing}. 
However, creating accurate 3D models of articulated objects remains challenging due to the intricate kinematic relationships among their constituent parts. 
Traditional modeling approaches typically involve an extensive manual effort, rendering them time-consuming and expensive~\cite{wang2019shape2motion, mu2021sdf}.
Recent methods leveraging neural rendering have significantly enhanced modeling efficiency by inferring 3D representations directly from visual observations~\cite{mildenhall2021nerf, liu2023paris, weng2024neural, wei2022self}.
Nevertheless, these methods still suffer from scalability limitations, as they often require dense multi-view imagery and considerable user interaction, restricting their practicality for large-scale generation of articulated objects~\cite{zhang2021learning, wang2024sm, wu2025reartgs, guo2025articulatedgs}.

Recent generative methods have demonstrated promise for synthesizing articulated 3D objects through diffusion-based frameworks~\cite{lei2023nap, liu2024cage, liu2024singapo, wu2025dipo, su2025artformer}. 
NAP~\cite{lei2023nap} was the first to employ a diffusion model for articulated object generation, explicitly representing articulated objects as tree-structured graphs, where nodes correspond to object parts and edges represent local kinematic relations. 
Building upon this representation, CAGE~\cite{liu2024cage} explicitly introduced adjacency graphs to guide the generative process, enhancing controllability and ensuring the generated assets conform to user-defined structural constraints. 
Further development~\cite{liu2024singapo, wu2025dipo, su2025artformer}, such as SINGAPO and ArtFormer, introduced semantic conditions derived from images or text prompts as high-level guidance for the generative pipeline. 
Specifically, SINGAPO leveraged single-view images to condition generation, while ArtFormer employed transformer-based diffusion architectures conditioned on textual inputs, thus facilitating more intuitive and diverse conditional generation.

However, existing generative approaches predominantly employ graph-based or tree-structured representations that emphasize local adjacency relationships between connected parts, neglecting the explicit modeling of global kinematic coherence across arbitrary articulation states. 
As a result, several critical challenges persist in current models:
1) the inference of joint types and motion directions remains unstable due to reliance on purely local context, 
often causing ambiguity or incorrect estimations of the kinematic configuration;
2) inadequate consideration of global semantic and spatial relationships between parts results in inconsistent or misaligned assemblies, 
where parts appear individually plausible but collectively lack structural coherence;
3) current methods generally fail to explicitly capture continuous articulation dynamics across multiple states, 
leading to physically implausible behaviors such as discontinuous motion trajectories, axis drift, or part interpenetration. 
Consequently, the primary limitation stems not from the graph-based representations themselves, but from their insufficient capability to incorporate global articulation consistency and cross-state kinematic reasoning.

To address these challenges, we propose ArtGen, a conditional diffusion framework for generating articulated 3D objects from a single-view image or text descriptions depicting an object in any articulation state. 
To ensure consistent articulation across an object’s full range of motion during training, 
we introduce a cross-state learning scheme that employs Monte Carlo sampling over continuous joint motion intervals, effectively exposing the model to a broad distribution of articulation states. 
We further employ a Chain-of-Thought reasoning strategy to infer robust structural priors from the input image or text, explicitly modeling part semantics, joint types, and connectivity. 
These inferred structural priors guide the generation process, thereby mitigating ambiguities in local kinematic predictions and enhancing global semantic and spatial coherence. 
To accurately capture diverse kinematic relationships and part interactions, we integrate a sparse expert diffusion Transformer~(DiT-MoE) into our framework. 
In this module, semantic and joint-type embeddings dynamically route features to specialized experts, enabling precise prediction of motion parameters and joint structures while ensuring kinematically coherent interactions among parts. 
Finally, the proposed framework can optionally incorporate a compositional 3D-VAE latent shape prior, enabling high-quality geometry synthesis without relying on static retrieval libraries.

Our contributions are summarized as follows:
\begin{itemize} [itemsep=1pt,topsep=1pt,parsep=1pt,leftmargin=20pt]
    \item We propose a conditional diffusion framework that explicitly enforces articulation consistency across arbitrary part-level states by cross-state learning, 
    improving the reliability of motion prediction and reducing structural-motion entanglement.
    
    \item We introduce a structured reasoning and generation architecture that combines Chain-of-Thought inference with a sparse expert diffusion transformer. 
    This design enables accurate prediction of part semantics, joint types, and connectivity, while supporting specialized modeling of diverse kinematic patterns.
    
    \item Extensive experiments on the PartNet-Mobility benchmark demonstrate that our method achieves superior reconstruction quality and controllability compared to existing approaches.
\end{itemize} 

%% file: figures/fig1_overview.tex
\begin{figure*}[!ht]
    \centering
    \includegraphics[width=0.98\linewidth]{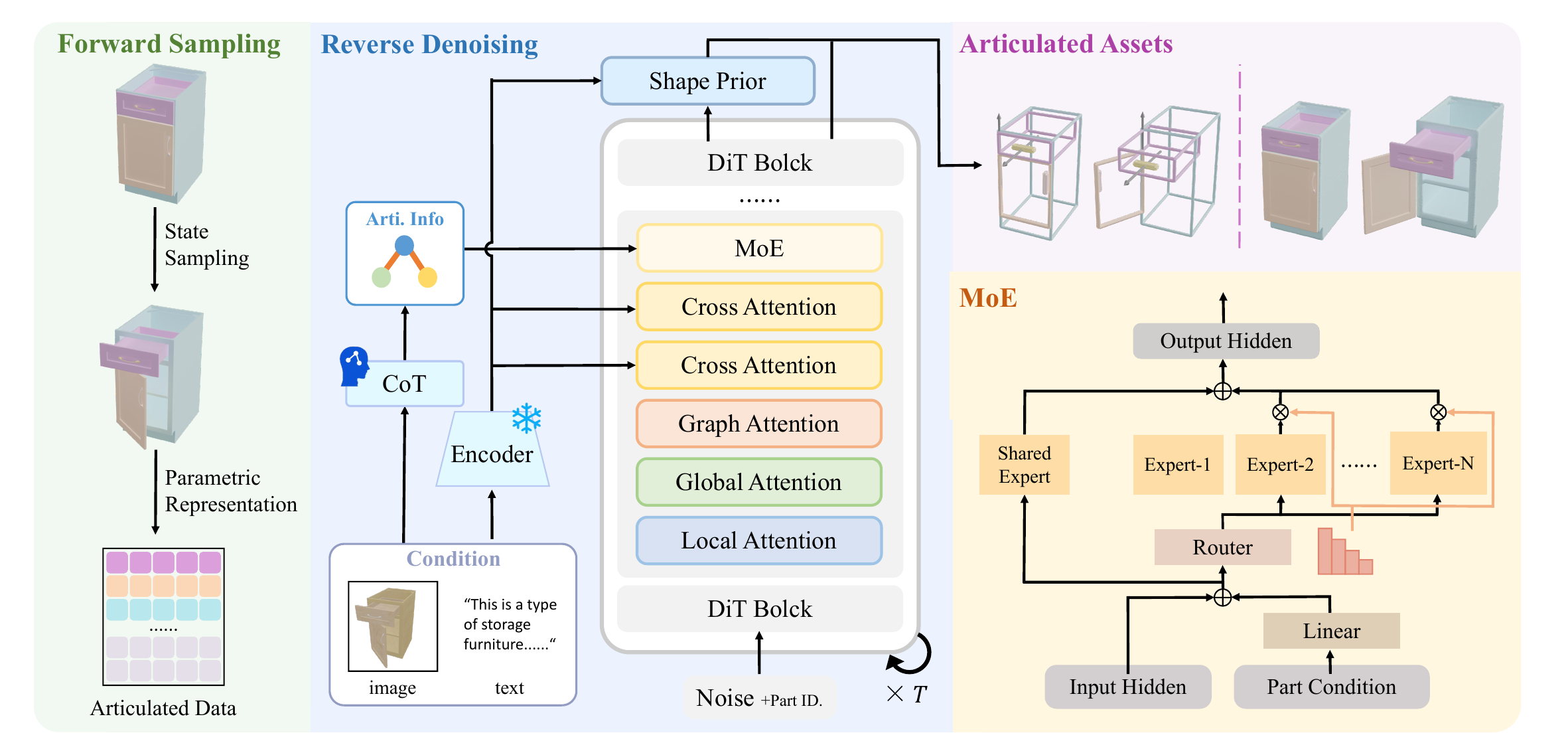}
    \caption{
        \textbf{Overview of the \methname\ pipeline.}
        Given an image or text description as input, our method generates both the geometric and kinematic attributes of articulated assets. We design a conditional diffusion framework that generates articulated 3D objects by leveraging cross-state learning, Chain-of-Thought reasoning, and a sparse expert diffusion transformer, ensuring consistent articulation and accurate kinematic modeling across arbitrary states.
  }
  \label{fig:fig1_overview}
\end{figure*}

%% file: sec/2_related_work.tex
\section{Related Work}
\label{sec:related_work}

\paragraph{Articulated Object Generation.}
Recent research on articulated object generation primarily focuses on reconstructing 3D kinematic structures using generative models. 
NAP~\cite{lei2023nap} first formulates articulated objects as graph-structured representations and performs unconditional articulated 3D object generation. 
CAGE~\cite{liu2024cage} enables constraint-guided controllable generation, but its conditioning remains coarse-grained. 
Building upon this, SINGAPO~\cite{liu2024singapo} introduces a diffusion model conditioned on a “rest” state image, while DIPO~\cite{wu2025dipo} further incorporates paired observations from two states to guide articulated asset synthesis.
However, ~\cite{liu2024cage, liu2024singapo, wu2025dipo} do not actually generate geometry. Instead, they retrieve shapes from a pre-existing dataset, which fundamentally limits their ability to produce novel and diverse articulated structures. 
ArtFormer~\cite{su2025artformer} addresses this by jointly generating geometry and kinematics from text or image inputs. 
Meanwhile, DreamArt~\cite{lu2025dreamart} synthesizes part-level geometry and motion by leveraging joint-aware video composition from a single input image.
However, the controllability of existing methods remains limited, primarily due to the lack of explicit modeling of articulation dynamics. The proposed \methname\ addresses this issue by performing part-level state sampling, enabling the model to learn consistent articulation behavior across different motion states.

\paragraph{3D Shape Generation.}
Previous studies on 3D object generation have explored a wide range of representations, including voxels~\cite{choy20163d, smith2017improved, xie2018learning}, point clouds~\cite{achlioptas2018learning, luo2021diffusion}, signed distance fields (SDFs)~\cite{chen2019learning, cheng2023sdfusion, li2023diffusion}, neural radiance fields (NeRFs)~\cite{mildenhall2021nerf, cao2023large, xu2023dmv3d}, and 3D Gaussian splatting (3DGS)~\cite{zhang2024gaussiancube, lin2025diffsplat}. 
While these approaches achieve impressive results in generating single holistic objects, they typically treat a 3D object as a monolithic entity and overlook the decomposition into constituent parts.
More recent efforts~\cite{liu2024part123, chen2025partgen} have attempted to generate NeRF~\cite{mildenhall2021nerf} or NeuS~\cite{wang2021neus} representations with part annotations by integrating multi-view diffusion or 2D segmentation models~\cite{kirillov2023segment, ravi2024sam}. 
Although effective, these approaches heavily rely on segmentation quality. 
In contrast, PartCrafter~\cite{lin2025partcrafter} directly synthesizes structured, part-level 3D objects without requiring any 2D or 3D segmentation supervision.
Our method introduces a local–global attention mechanism that jointly models fine-grained local geometric details and global structural relationships at the part level, enabling high-quality part-level geometry generation and structure-consistent articulated object modeling.

%% file: sec/3_methods.tex
\section{Method}
\label{sec:method}

This section outlines the specific framework of \methname\ as shown in Fig.~\ref{fig:fig1_overview}. 
We first introduce the representation of articulation data, providing a unified data representation for different articulated objects (Sec.~\ref{method:art_rep}). 
Next, to better generate high-quality geometric shapes while modeling the kinematic parameters, we learn a composite shape prior (Sec.~\ref{method:shape_prior}). 
Finally, building upon this, given the conditional information from an image or text description, we utilize Chain-of-Thought (CoT) to extract additional articulated object information and generate articulated assets using a Diffusion Transformer enhanced with Mixture-of-Experts (MoE) (Sec.~\ref{method:art_dit}).

\subsection{Articulation Data Representation}
\label{method:art_rep}
\input{sec/methods/art_rep}

\input{figures/fig2_shape_prior}

\subsection{Part-level Shape Prior}
\label{method:shape_prior}
\input{sec/methods/shape_prior}

\subsection{Articulation Diffusion Transformer}
\label{method:art_dit}
\input{sec/methods/art_dit}

%% file: sec/methods/art_rep.tex
We parameterize the joint attributes of each part in the articulated asset using both geometric and kinematic attributes $a
_i = \{ b_{i}, f_{i}, l_{i}, r_{i}, s_{i} \}$, with the parameterization method being similar to that used in CAGE~\cite{liu2024cage}.
The geometric parameters for the $i$-th part are defined by the following two attributes:

\begin{itemize}[leftmargin=*]
    \item \textbf{Bounding box, $ b_{i} \in \mathbb{R}^{9} $:} Unlike previous methods that normalize all objects to a "rest" state, we account for the various states of the objects. As a result, the Axis-Aligned Bounding Box (AABB) in a canonical orientation fails to accurately represent the orientation of objects with some joint types (e.g., rotational joints) across different states. Instead, we use an Oriented Bounding Box (OBB) to align the object to a canonical pose.
    
    \item \textbf{Shape latent code, $ f_{i} \in \mathbb{R}^{F} $:} We utilize a pre-trained 3D Variational Autoencoder (VAE) model, where the 3D point cloud of each part is processed through a geometry encoder to generate a shape latent code that effectively captures the detailed geometric information of the part.
    
\end{itemize}

The kinematic parameters for the $i$-th part and its parent part are composed of the following two attributes:

\begin{itemize}[leftmargin=*]
    \item  \textbf{Joint axis, $ l_{i} \in \mathbb{R}^{6} $:} The joint axis defines the direction and type of relative motion between components, consisting of the origin coordinates and direction vector. The origin coordinates specify th e axis's position in 3D space, while the direction vector indicates the rotation or translation direction of the joint.
    
    \item  \textbf{Joint range, $ r_{i} \in \mathbb{R}^{4} $:} We represent each joint’s motion relative to its parent component using the joint axis, which defines the minimum and maximum positions it can achieve during motion. For revolute/continuous joints, the first two parameters specify the lower and upper bounds of the rotational angle. For prismatic/screw joints, the last two parameters define the minimum and maximum translation distances.
    
\end{itemize}

The kinematic state parameters for the $i$-th part are composed of the following attributes:

\begin{itemize}[leftmargin=*]
    \item  \textbf{Part State, $ s_{i} \in \mathbb{R}^{1} $:} This attribute characterizes the degree of joint motion, with values normalized within the joint range $ [0, 1]$. A value of $0$ corresponds to the joint's initial or closed state, while $1$ indicates the joint is fully open or at its maximum motion state. Intermediate values represent the intermediate states between these two extremes.
    
\end{itemize}

%% file: figures/fig2_shape_prior.tex
\begin{figure}[!t]
    \centering
    \includegraphics[width=0.99\linewidth]{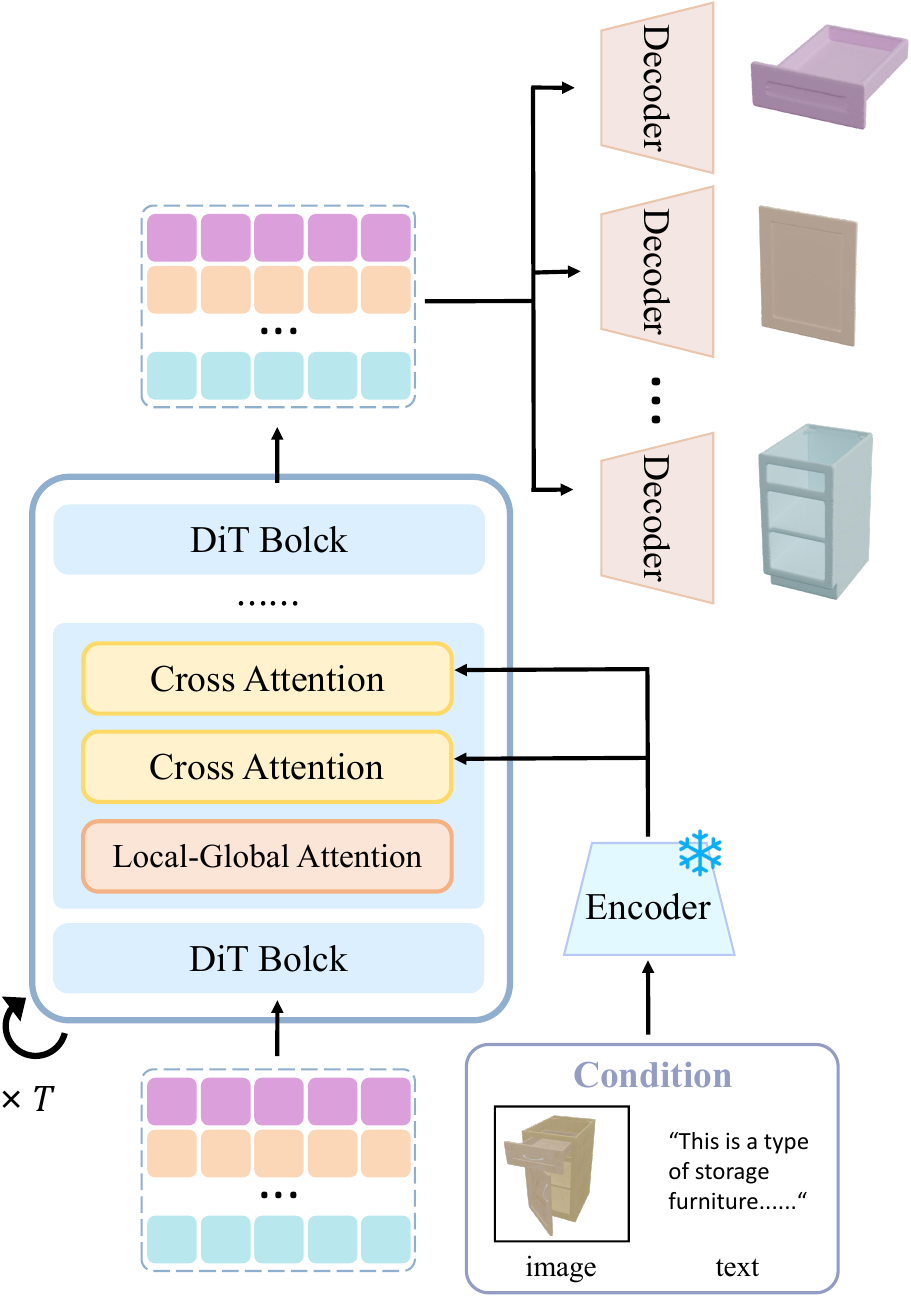}
    \caption{
        The \textbf{Part-Level Shape Prior} with a local-global attention mechanism enables part-level shape generation by jointly modeling fine-grained local geometry and global structural relationships among articulated parts.
  }
  \label{fig:fig2_shape_prior}
\end{figure}

%% file: sec/methods/shape_prior.tex
Simultaneously modeling and generating high-quality geometric shapes and kinematic relationships is not an easy thing
Therefore, we first learn the part-level shape priors, similar to the approach used in PartCrafter~\cite{lin2025partcrafter}, as illustrated in Fig.~\ref{fig:fig2_shape_prior}.

\paragraph{Shape Prior.}
An articulated object consists of multiple sub-parts, each with interrelated shapes. Our approach is to use multiple sets of shape latent codes for compositional shape generation, where each shape latent code is responsible for a single sub-part of the articulated object. 
Specifically, we first employ a 3D Variational Autoencoder (VAE) to encode the 3D part shapes of the articulated object into a set of latent vectors $ z =\{ \text{z}_{i}\}_{i=1}^{N} \in \mathbb{R}^{N \times F}$, where $N$ is the number of parts in the articulated object and $F$ is the dimensionality of each latent code.
To distinguish between different parts, we add learnable part identity embedding $ e_{i} \in \mathbb{R}^{E} $  to each part. 
Subsequently, we introduce geometric shape conditioning and part-level guidance for joint generation. We train a conditional diffusion model $\epsilon(z_t, t, c_S, c_I)$, where $c_S=E_{s}(x)$ represents the geometric shape encoding features, and $c_I=E_{I}(x)$ denotes the encoding features conditioned on the image or text description.

\paragraph{Local-Global Attention.}
To better capture the features of the parts, we incorporate a local-global attention mechanism into the compositional shape prior model. Specifically, we independently apply local attention to the latent encoding of each part $\mathbf{z}_i$, to capture the local features within each part, ensuring that their internal structures remain distinct. After capturing the part-level features, we then employ global attention to model the interactions between parts.
\begin{equation}
    \begin{aligned}
        \text{Attn}_{i}^{\text{local}} &= \text{softmax}(\text{z}_{i}\text{z}_{i}^{T}) \in \mathbb{R}^{F \times F}, \\
        \text{Attn}^{\text{global}} &= \text{softmax}(zz^{T}) \in \mathbb{R}^{NF \times NF},
    \end{aligned}
\end{equation}
where $\text{Attn}$ represents the generated attention map. The effective exchange of information between the local and global attention mechanisms enables fine-grained feature capture while also incorporating global context.

%% file: sec/methods/art_dit.tex
\paragraph{Multimodal Conditioning.}
Leveraging the flexible conditioning mechanism provided by diffusion models~\cite{wang2025diffusion}, we incorporate multiple conditional input modalities simultaneously through a task-specific encoder $E_{\phi}$ and cross-attention modules.
For image-conditioned generation, we directly use DINOV3~\cite{simeoni2025dinov3} to extract patchified features from the image, which are then injected using the cross-attention mechanism. The resulting features are combined with the original input and passed to the next stage. The process is defined as:
\begin{equation}
     \hiddenfeat =  \hiddenfeat + \text{CrossAttn}(\text{Norm}(\hiddenfeat), I_{feature}),
\end{equation}
We generate paired data using the following procedure: (1) sample all articulated objects from the preprocessed dataset, and (2) render images from different views using Blende~\cite{blender}.

For text-conditioned generation, the text input is processed by a pre-trained text encoder~\cite{colin2020exploring}, generating a sequence of conditioning tokens. These tokens are then integrated into the transformer via cross-attention layers. The process is defined as:
\begin{equation}
    \hiddenfeat =  \hiddenfeat + \text{CrossAttn}(\text{Norm}(\hiddenfeat), T_{feature}),
\end{equation}
The text data is generated based on the image sampling process, where GPT-4o is used to generate textual descriptions of the articulated objects from multiple views.
Further details are provided in the supplementary materials.

\paragraph{Graph Analysis via CoT.}
By incorporating Chain-of-Thought (CoT) techniques, we leverage large vision-language models to predict the articulated connectivity graph, joint types, and part semantics from images or text descriptions, thereby providing structural priors for the diffusion process. Specifically, we utilize GPT-4o to analyze the input through the following steps:
\begin{enumerate*}[label=(\arabic*), itemjoin={\space}]
    \item Analyzing the prompt input to identify the number and labels of candidate parts and estimate their approximate spatial relationships;
    \item Combining part morphological features with spatial relationships to infer interactions or relative motions between parts;
    \item Inferring the connectivity between parts to generate the connectivity graph, joint types, and part semantics. 
\end{enumerate*}

Subsequently, we convert the predicted joint connectivity graph into an adjacency matrix, which is used as an attention mask to guide the diffusion model’s attention computation over valid structural connections. The joint types and part semantics are encoded to guide the Mixture-of-Experts model in dynamic routing selection. For more details, please refer to the supplementary materials.

\paragraph{Transformer with MoE.}
\label{para:moe} 
Our model is built upon the DiT framework~\cite{peebles2023scalable}, which includes skip-connections, RMSNorm~\cite{zhang2019root}, and the injection of global, local, and graph features. 
To enhance the modeling capacity while maintaining computational efficiency, we adopt the DiT-MoE approach~\cite{fei2024scaling}, replacing the standard feedforward module in each block with a Mixture-of-Experts (MoE) mechanism, inspired by the successful application of MoE architectures in large language models.
We integrate $N$ parallel experts into the output projection layer of the attention block, where a routing classifier selects experts based on joint type and part semantic embeddings, while maintaining a shared expert. The MoE output is formulated as follows:
\begin{equation}
    MoE(\hiddenfeat) = SE(\hiddenfeat) + \sum_{i=1}^{N}{G(\hiddenfeat) \cdot E_{i}(\hiddenfeat) },
\end{equation}
where $\hiddenfeat$ is the input to the layer, $ SE $ denotes the function computed by the shared expert, $ E_{i}$ denotes the function computed by expert $i$, and $ G_{i}$ is the routing function that determines the input-conditioned weights for the experts. We utilize a sparse MoE setup, selecting the top-$k$ experts, where $G_{i}(\hiddenfeat)=\text{softmax}_{i}({\text{TopK}(g(\hiddenfeat), k)})$, with $\text{TopK}( \cdot, k)$ retaining the top-$k$ entries and assigning the remaining entries a value of $-\infty$.
This design enables the model to specialize across different joint types and part semantics, thereby improving performance with minimal overhead.

\paragraph{Training Objective.} 
Our model follows the training process of a denoising diffusion probabilistic model~\cite{ho2020denoising}. Specifically, given an articulated part with multiple attributes  $\mathcal{A}_{0} = \{ a_i \}_{i=1}^{N}$ and a condition $c$, we perturb the latent representation by adding Gaussian noise $ \epsilon \sim \mathcal{N}(0, \mathbf{I}) $ at a noise level $t$, resulting in a noisy latent representation $ \mathcal{A}_{t}=t\mathcal{A}_{0}+(1-t)\epsilon $.
The model is then trained to predict the noise $\epsilon$ given the noisy latent $\mathcal{A}_{t}$ at the noise level $t$, conditioned on the condition single-view image (or a text description) $\mathbf{c}$, by minimizing the following objective:
\begin{equation}
    \mathcal{L} =\mathbb E_{\mathcal{A}, \epsilon, t}
    \left[ \left\|
    \epsilon - \epsilon_{\theta}(\mathcal{A}_{t}, t, \mathbf{c})
    \right\|^2 \right],
\end{equation}
where $\epsilon_{\theta}$ is a learnable network.

%% file: tables/tab1_sota.tex
\begin{table*}[t!]
\renewcommand\arraystretch{1.1}
    \centering
    \caption{Quantitative comparison of generation quality for articulated objects. The left part of the table presents the part meshes reconstructed from the generated shape encodings, while the right part displays the retrieved part meshes.}
    \begin{tabular}{@{}ccccccccc@{}}
        \toprule
        
        \multirow{2}{*}{\begin{tabular}[c]{@{}c@{}}Method\end{tabular}} 
        & \multicolumn{4}{c}{Part SDF Shape} & \multicolumn{4}{c}{Part Retrieval Shape} \\ 
        \cmidrule(lr){2-5} \cmidrule(l){6-9}  
        & POR $\downarrow$ & MMD $\downarrow$ & COV $\uparrow$ & 1-NNA $\downarrow$ 
        & POR $\downarrow$ & MMD $\downarrow$ & COV $\uparrow$ & 1-NNA $\downarrow$ \\ 
        
        \midrule
        NAP & 3.648 & 0.076 & 0.171 & 0.967 & 1.311 & 0.021 & 0.540 & 0.565 \\
        CAGE & - & - & - & -  & 1.398 & 0.031 & 0.497 & 0.565 \\
        
        SINGAPO & - & - & - & -  & 0.594 & 0.021 & 0.448 & 0.413 \\
        ArtFormer & 0.911 & 0.048 & 0.612 & 0.691 & 0.253 & \textbf{0.014} & 0.543 & 0.396 \\
        Ours & \textbf{0.694} & \textbf{0.020} & \textbf{0.564} & \textbf{0.538} & \textbf{0.189} & 0.018 & \textbf{0.551} & \textbf{0.311} \\
        
        \bottomrule
    \end{tabular}
    \label{tab:tab1_sota}
\end{table*}

%% file: figures/fig3_sota.tex
\begin{figure*}[!ht]
    \centering
    \includegraphics[width=0.99\linewidth]{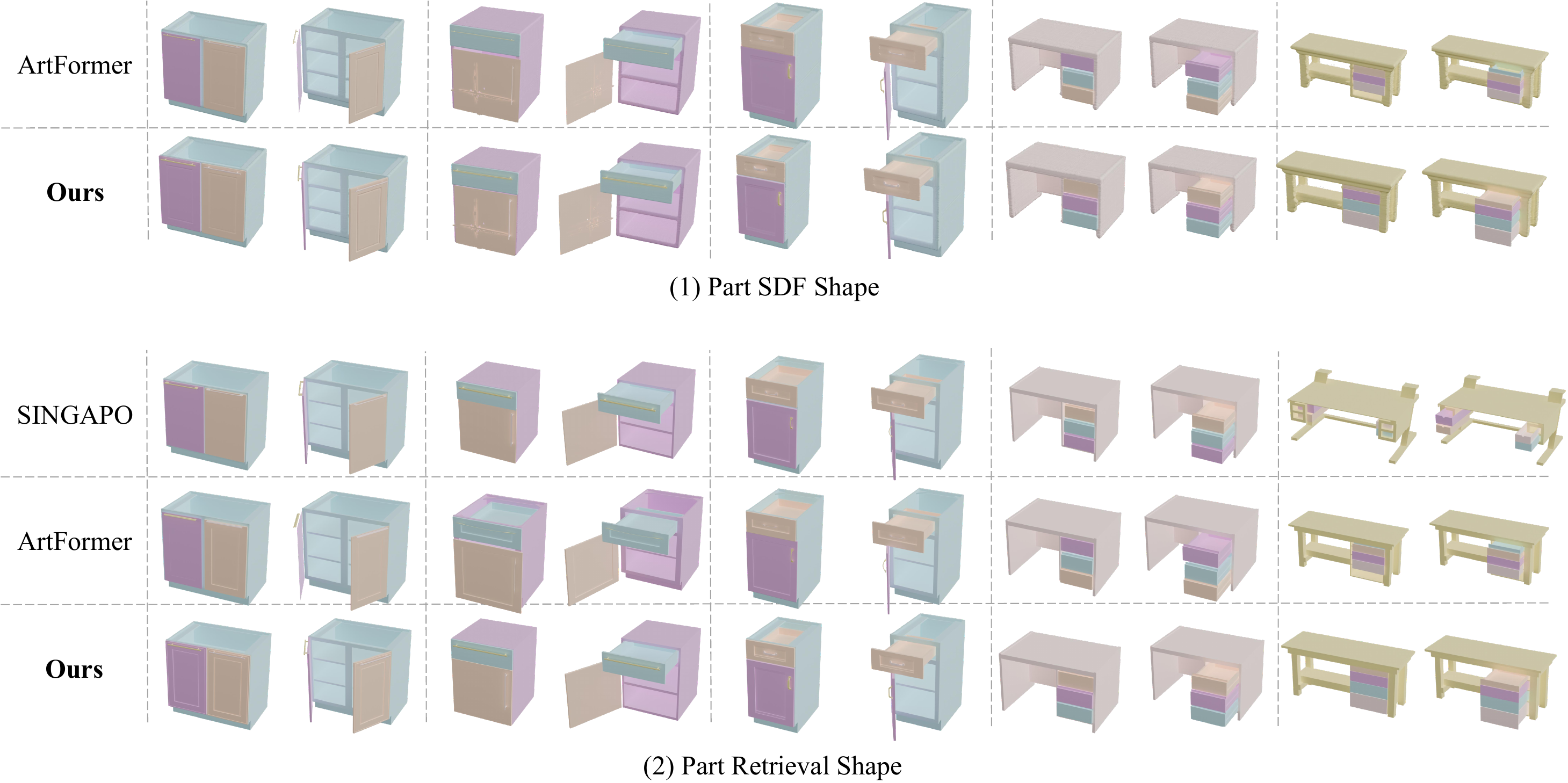}
    \caption{
        Qualitative comparison of generation quality for articulated objects. 
  }
  \label{fig:fig3_sota}
\end{figure*} 

%% file: sec/4_experiments.tex
\section{Experiments}
\label{sec:exp}

\subsection{Experimental Setup}
\label{exp:exp_set}
\input{sec/experiments/exp_set}

\subsection{Results}
\label{exp:compare_sota}
\input{sec/experiments/compare_sota}

\input{tables/tab3_abla}

\subsection{Ablations}
\label{exp:abla}
\input{sec/experiments/abla}

%% file: sec/experiments/exp_set.tex
\paragraph{Datasets.} We train \methname\ and other baseline models on 10 categories from the PartNet-Mobility~\cite{xiangsapien} dataset (Dishwasher, Laptop, Microwave, Oven, Refrigerator, Suitcase, Table, TrashCan, Washer, and Storage), extracting both geometric shapes and kinematic information from the dataset assets while normalizing the coordinate system. For each articulated object instance in the dataset, we render high-resolution images using Blender and generate corresponding textual descriptions using GPT-4o. 

\paragraph{Baselines.} We compare \methname\ with the recent SOTA methods. NAP~\cite{lei2023nap} is a prior work on unconditional generation of articulated objects, using a pre-trained occupancy shape autoencoder to generate part geometries. In contrast, CAGE~\cite{liu2024cage} retrieves part shapes from the dataset instead of generating them. SINGAPO~\cite{liu2024singapo} generates articulated part attributes from a single-view input and assembles objects through a part retrieval method similar to that of CAGE. ArtFormer~\cite{su2025artformer}, on the other hand, generates articulated object tree structures conditioned on both text and image inputs.

\paragraph{Metrics.} We follow the Instantiation Distance (ID) evaluation metric proposed by NAP to assess the accuracy of the model generation. The ID considers both geometric shapes and kinematic information of articulated objects by sampling various joint states. Thus, a smaller ID value between two articulated objects indicates higher similarity, and vise versa. It is defined as:
\begin{align}
\begin{split}
    ID(O_1, O_2) \approx &\frac{1}{M}\sum_{q_1 \in Q_1} \left[ 
    \min_{q_2\in Q_2}\left(\tilde d(O_1, q_1, O_2, q_2)\right)
    \right] \\
    +
    &\frac{1}{M}\sum_{q_2 \in Q_2} \left[ 
    \min_{q_1\in Q_1}\left(\tilde d(O_1, q_1, O_2, q_2)\right)
    \right],
\end{split}
\end{align}
where $Q$ is a set of $M$ uniformly sampled joint poses from the range of joint states, and $\tilde d$ is the minimum chamfer distance calculated across all possible normalized orientations of the meshes.The following evaluation metrics are defined:
\begin{enumerate*}[label=(\arabic*), itemjoin={\space}]
    \item Minimum Matching Distance (MMD) describes the average minimum distance between the generated objects and ground truth matches;
    \item Coverage (COV) represents the ratio of ground truth instances that have corresponding matches in the generated objects;
    \item Nearest Neighbor Accuracy (1-NNA) measures the distance between the generated objects and ground truth objects after performing 1-NN clustering. 
\end{enumerate*}
Additionally, we use Part Overlapping Rate (POR) to evaluate the degree of interpenetration between sub-parts.

\paragraph{Implementation Details.} All of our model training is conducted on multiple NVIDIA RTX 4090 devices. 
We use the DDPMScheduler as the diffusion scheduler, with a total of 1000 training time steps, employing a linear beta schedule to control the noise addition process. 
To support the inherent permutation invariance of the parts, we shuffle the order of the parts during training. 
Additional details on the experimental setup can be found in the Supplementary Materials.

%% file: sec/experiments/compare_sota.tex
\paragraph{Generation Quality.}
Fig.~\ref{fig:fig3_sota} and Tab.~\ref{tab:tab1_sota} report the quantitative and qualitative results on the PartNet-Mobility dataset. 
Since CAGE and SINGAPO cannot directly generate geometry, we compare two groups: the first group includes NAP, ArtFormer, and our model, while the second group consists of all models that retrieve suitable shapes from the dataset to assemble articulated objects. 
In the first group, our model outperforms all other methods across all metrics, generating more realistic and coherent articulated objects. 
In the second group, although ArtFormer achieves better MMD, indicating better retrieval-generated accuracy, our model excels in other metrics, producing more realistic and structurally consistent articulated objects.
More details are provided in the supplementary material.

\input{figures/fig4_image_gen}
\paragraph{Image Guided Generation.}
Fig.~\ref{fig:fig4_image_gen} shows the results of image-guided generation. 
The model generates articulated objects conditioned on input images, demonstrating its ability to accurately capture geometric and kinematic properties based on visual cues. 
This highlights the model's capacity to generate diverse and structurally consistent objects based on image input.

\input{figures/fig5_txt_gen}
\paragraph{Text Guided Generation.}
Fig.~\ref{fig:fig5_txt_gen} presents the results of text-guided generation. 
The model generates articulated objects conditioned on textual descriptions, with the generated objects aligning well with the provided descriptions. 
This showcases the model’s ability to translate semantic information into geometrically and kinematically coherent 3D objects.

\input{tables/tab2_user_sota}

\paragraph{User Study Evaluation.}
To further evaluate the plausibility of the generated articulated assets, we conducted a user study focusing on the consistency between the generated objects and their corresponding input conditions. 
Specifically, 20 participants were presented with 20 different input conditions and the corresponding generated hinge objects. 
Each participant rated the alignment between the generated objects and the input conditions from three perspectives: geometric shape, kinematic relationships, and overall consistency. 
The ratings were given on a five-point Likert scale, ranging from 1 as the worst to 5 as the best, reflecting the alignment with the input conditions as well as the rationality of the object’s joints. 
We collected all participants’ evaluations and summarized the aggregated results in Tab.~\ref{tab:tab2_user_sota}.
The results in the table indicate that participants consistently perceived the articulated assets generated by our model as having more accurate geometry, more physically plausible motion, and greater overall coherence compared to existing methods.

%% file: figures/fig4_image_gen.tex
\begin{figure}[!h]
    \centering
    \includegraphics[width=0.79\linewidth]{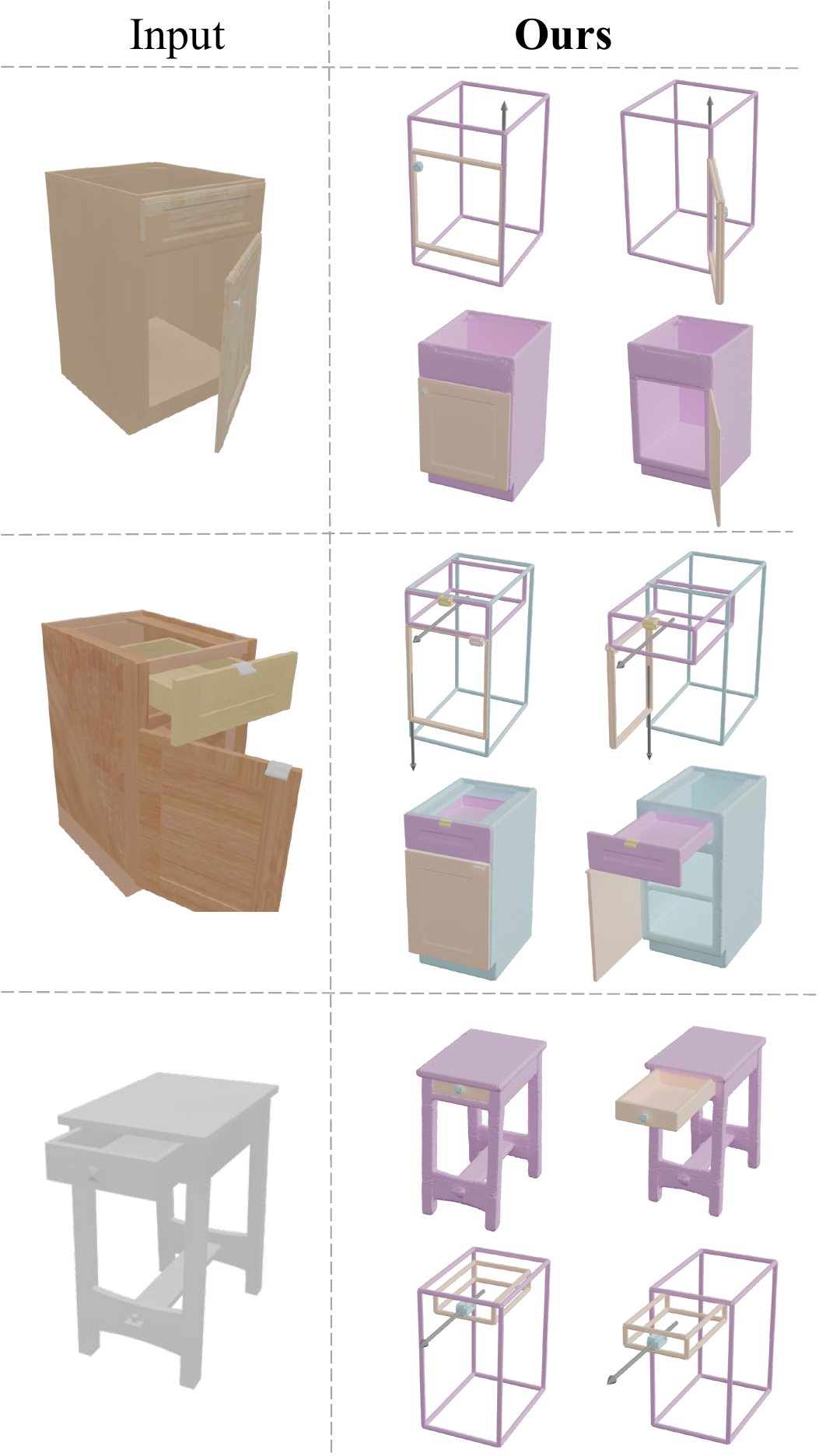}
    \caption{
    Illustration of our model's input and output: an input image (left) and the generated articulated object (right).
  }
  \label{fig:fig4_image_gen}
\end{figure} 

%% file: figures/fig5_txt_gen.tex
\begin{figure}[!h]
    \centering
    \includegraphics[width=0.89\linewidth]{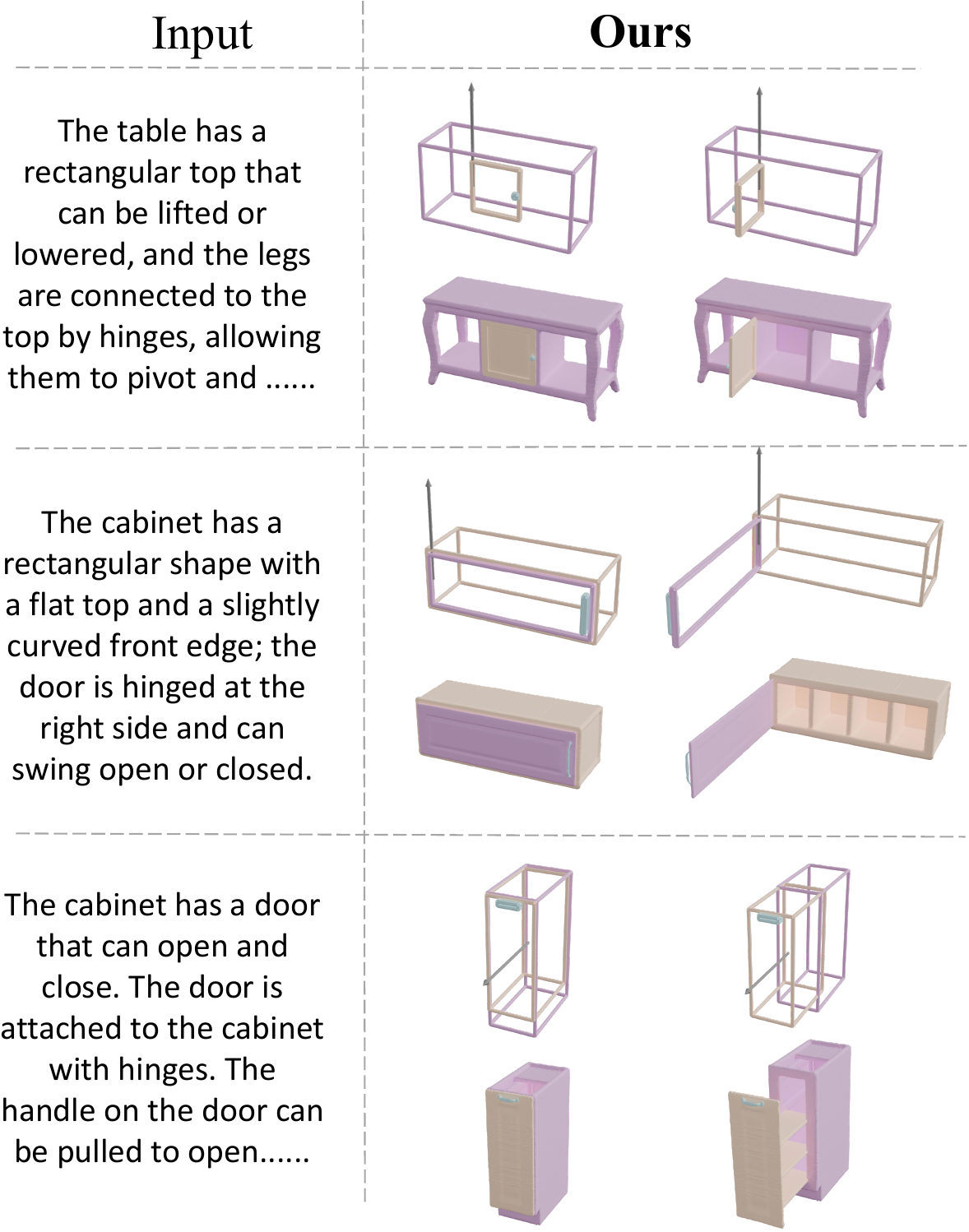}
    \caption{
        Illustration of our model's input and output: an input text description (left) and the generated articulated object (right).
  }
  \label{fig:fig5_txt_gen}
\end{figure} 

%% file: tables/tab2_user_sota.tex
\begin{table}[h]
    \caption{User Evaluatation.}
    \label{tab:tab2_user_sota}
    \centering
    \begin{tabular}{@{}cccc@{}}
        \toprule
        Method 
        & Geo.$\uparrow$   & Kin$\uparrow$ & Pref.$\uparrow$   \\ 
        
        \midrule
        NAP      & 1.95 & 2.50 & 2.35  \\
        CAGE     & 3.20 & 3.30 & 3.15 \\
        SINGAPO  & 3.55 & 3.35 & 3.45 \\
        ArtFormer & 3.95 & 3.15 & 3.70 \\
        Ours     & \textbf{4.05} & \textbf{4.20} & \textbf{4.15} \\
        \bottomrule
    \end{tabular}
\end{table}

%% file: tables/tab3_abla.tex
\begin{table}[h]
    \caption{
        Ablation Study on the Quality of Generated Objects.
    }
    \label{tab:abla}
    \centering
    \begin{tabular}{@{}lcccc@{}}
        \toprule
        &  \multirow{2}{*}{\begin{tabular}[c]{@{}c@{}}POR $\downarrow$\\ \text{${\times 10^{-2}}$}\end{tabular}} 
        & \multicolumn{3}{c}{ID}  \\ \cmidrule(lr){3-5}
        & & MMD$\downarrow$   & COV$\uparrow$   & 1-NNA$\downarrow$ \\ 
        
        \midrule
        Full & \textbf{0.694} & \textbf{0.020} & \textbf{0.564} & \textbf{0.538} \\
        w/o SP   & 2.535 & 0.043 & 0.426 & 0.824 \\
        w/o GA   & 0.703 & 0.024 & 0.526 & 0.552 \\
        w/o MoE  & 0.725 & 0.026 & 0.523 & 0.574 \\
        
        \bottomrule
    \end{tabular}
\end{table}

%% file: sec/experiments/abla.tex
To evaluate the effectiveness of our framework, we conduct ablation studies on the shape prior (SP), graph analysis (GA), and Mixture-of-Experts models (MoE).
Tab.~\ref{tab:abla}.
Removing the part-level shape prior forces the ArtGen diffusion Transformer to directly generate part geometries, limiting its ability to capture inter-part relationships and making it harder to produce coherent shapes. Without the Graph Analysis (GA) component, performance degrades slightly, as the model loses global understanding of part relationships, causing misalignments and inconsistencies. Adding the Mixture-of-Experts (MoE) component improves performance across all metrics, allowing the model to specialize in different joint motions and structures, thus enhancing kinematic accuracy and flexibility.

%% file: sec/5_conclusion.tex
\section{Conclusion}
\label{sec:conclusion}

In this paper, we present \methname\, a unified conditional diffusion framework for generating articulated 3D objects with geometric and motion fidelity from a single-view image or text description.
\methname\ explicitly models cross-state joint consistency through part-level Monte Carlo state sampling, enabling the generation of kinematically plausible and coherent assets across different motion states. 
The framework integrates a chain-of-thought reasoning module with a sparse expert diffusion transformer to infer semantic parts, joint types, and connectivity graphs, while maintaining high-quality geometry through a part-level shape prior. 
Extensive experiments on the PartNet-Mobility dataset demonstrate that ArtGen produces high-quality articulated objects under both image and text conditions, outperforming state-of-the-art methods in terms of geometric realism and articulation consistency.

\mypara{Limitations and Future Work.}
\begin{enumerate*}[label=(\arabic*), itemjoin={\space}]
    \item \methname\ focuses on motion consistency across joints but does not explicitly model physical dynamics such as joint friction, collisions, or gravity. As a result, some generated motions may be kinematically plausible but lack physical realism or feasibility. Future work will incorporate physical dynamics models to guide the generation of kinematically correct and physically plausible motions for articulated objects.
    \item The current framework primarily models geometric and kinematic features, neglecting appearance cues such as textures and materials in the input. This limitation leads to generated objects with unrealistic material properties. In the future, we plan to integrate texture and material priors to improve the visual realism of generated objects.
\end{enumerate*}